\documentclass{article}
\usepackage{amssymb}
\usepackage{amsmath}
\usepackage{amsthm}
\usepackage[a4paper]{geometry}
\usepackage{multirow}

\usepackage{algorithm}
\usepackage{algorithmic}
\usepackage[round]{natbib}
\usepackage{authblk}       

\usepackage{subfig}
\usepackage{graphicx}

\usepackage{tabularx}

\usepackage{booktabs}
\usepackage[table]{xcolor}

\definecolor{cvprblue}{rgb}{0.21,0.49,0.74}
\usepackage{hyperref}
\hypersetup{colorlinks,allcolors=cvprblue}

\newcommand\blfootnote[1]{%
  \begingroup
  \renewcommand\thefootnote{}\footnote{#1}%
  \addtocounter{footnote}{-1}%
  \endgroup
}

\title{Building a Mind Palace: Structuring Environment-Grounded Semantic Graphs for Effective Long Video Analysis with LLMs}

\author{
  Zeyi Huang$^{1\star}$, Yuyang Ji$^{3\star}$, Xiaofang Wang$^2$, Nikhil Mehta$^2$, Tong Xiao$^2$,\vspace{-0.8em}\\ Donghyun Lee$^1$,
  Sigmund Vanvalkenburgh$^1$, Shengxin Zha$^2$, Bolin Lai$^3$, \\ Yiqiu Ren, Licheng Yu$^2$, Ning Zhang$^2$,
  Yong Jae Lee$^{1\dagger}$, Miao Liu$^{2\dagger}$ \vspace{0.8em}
  \\
  {\hspace{-1.6cm}$^1$University of Wisconsin-Madison \hspace{0.9cm} $^2$Meta \hspace{0.9cm} $^3$UIUC}\\
}

\date{}

\begin{document}
\maketitle

\begin{abstract}
Long-form video understanding with Large Vision Language Models is challenged by the need to analyze temporally dispersed yet spatially concentrated key moments within limited context windows. In this work, we introduce VideoMindPalace, a new framework inspired by the ``Mind Palace", which organizes critical video moments into a topologically structured semantic graph. VideoMindPalace organizes key information through (i) hand-object tracking and interaction, (ii) clustered activity zones representing specific areas of recurring activities, and (iii) environment layout mapping, allowing natural language parsing by LLMs to provide grounded insights on spatio-temporal and 3D context. In addition, we propose the Video MindPalace  Benchmark (VMB), to assess human-like reasoning, including spatial localization, temporal reasoning, and layout-aware sequential understanding. Evaluated on VMB and established video QA datasets, including EgoSchema, NExT-QA, IntentQA, and the Active Memories Benchmark, VideoMindPalace demonstrates notable gains in spatio-temporal coherence and human-aligned reasoning, advancing long-form video analysis capabilities in VLMs.
\end{abstract}

\section{Introduction}
\label{sec:intro}
\blfootnote{$^\star$,$^\dagger$ equal contribution and advising. Correspondence to \{zeyihuang,yongjaelee\}@cs.wisc.edu, yuyangji@illinois.edu, miaoliu@meta.com}

One key challenge for long-form video understanding with Large Vision Language Models (VLMs) lies in jointly analyzing key moments that are temporally distant yet strongly correlated, as the context window of LLMs is bounded. Notably, one intriguing property of long-form video, which has been largely overlooked by existing long video VLM works~\cite{maaz2023video, lin2023video, li2024llava,zhang2023simple, wang2024videotree, fan2025videoagent, ren2024timechat, song2024moviechat, cheng2024videollama, xu2024slowfast, xu2024pllava, li2025llama, ma2023vista, maaz2024videogpt+}, is that these critical moments often occur within a limited set of \emph{spatial zones} despite their temporal dispersion. This unique combination of extensive temporal span and concentrated spatial activity zones calls for a novel representation tailored specifically for long-form video analysis.

Inspired by ``Mind Palace'', a memorization enhancement technique that involves associating temporal information with specific locations, then organizing these locations in a structured and orderly manner to enhance memory recall~\cite{yates2013art}, we introduce a novel VLM system -- VideoMindPalace, which captures a rich topological and hierarchically-organized semantic graph representation of an environment 
with the versatility to encode the necessary information required for tasks like spatial grounding, temporal grounding, 3D environment comprehension using natural language suitable for parsing by an LLM. 

\begin{figure}[t]
    \centering
    \includegraphics[width=0.85\columnwidth]{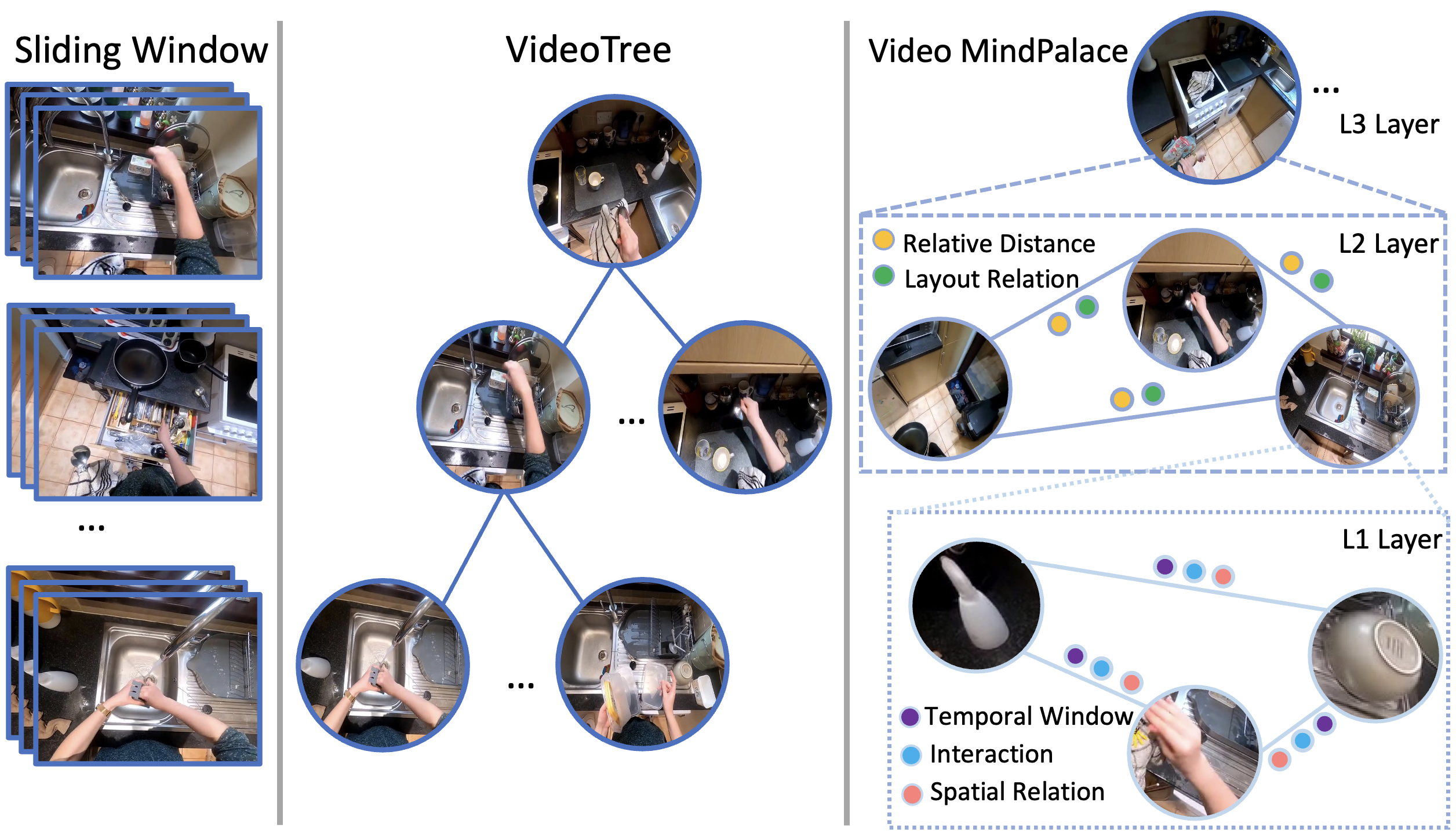}
    \vspace{-0.5em}
    \caption{(Right) Our VideoMindPalace represents video data as a layered, topological structured graph, where nodes capture spatial concepts (e.g., objects, activity zones, rooms), and edges signify spatiotemporal, layout relationships and human-object interaction. This graph can be represented in JSON format and used as input to text-only LLMs. (Center) VideoTree~\cite{wang2024videotree} extracts query-relevant information by organizing videos as tree structures, with deeper branches capturing finer, query-specific details. A captioner then generates video descriptions from this structure, enabling the LLM to perform reasoning over long videos. (Left) LLoVi~\cite{zhang2023simple} processes videos following temporal order, where visual captioners sequentially generate textual descriptions within each temporal sliding window, which the LLM then aggregates for reasoning.}
    \label{fig:teaser}
    \vspace{-1.5em}
\end{figure}

Specifically, VideoMindPalace comprises (i) a collection of human-object tracking, capturing consistent interaction between human and objects, as well as their spatial proximity; (ii) clustered activity zones, representing specific zones where human engages in activities during different temporal footprint; and (iii) the environment layout, reflecting the topological arrangement of the activity zones that human explored. Importantly, these components are constructed using visual perception models of human activity~\cite{zhang2022bytetrack, tang2024egotracks} and mapping~\cite{tschernezki2024epic} rather than a naive sampling of frames. Such models capture information like ``What did I do before/after using an object'', ``Where did I use an object?'', ``Is there an unobstructed path between two locations?'', etc., resulting in representations closely aligned with both activities and environmental context. 

The community has developed a wealth of methods for understanding video. One approach, Video LLMs~\cite{maaz2023video, lin2023video, li2024llava}, involves end-to-end models fine-tuned on large-scale labeled video datasets, achieving strong results of short videos. While our investigation is more relevant to a new direction~\cite{zhang2023simple, wang2024videoagent} that leveraged the long-sequence reasoning abilities of LLMs to address long-form video understanding without additional training. These methods generally use vision-language models to caption densely sampled frames, transforming the video into a text format that an LLM can analyze to respond to queries. While promising, this strategy faces a significant limitation: Information Overload. Long videos contain high redundancy, and current methods lack effective strategies to filter this out, often overwhelming the LLM with excessive irrelevant information, which leads to errors and inefficiency.

Recent approaches aim to organize long-form video into structured representation, such as search tree~\cite{wang2024videotree}, or temporal and object memory bank~\cite{fan2025videoagent}. In contrast, our VideoMindPalace system represents the video as a layered topologically structured graph, where nodes represent spatial concepts (e.g., objects, activity zones, room layouts) and edges represent human-object interaction, spatio-temporal, layout relationships. Our proposed hierarchical graph structure mimics human mind palace process, and thereby is an ideal representation of long-form video to enhance LLM comprehension and avoid the exhaustive search for different visual queries as in previous works~\cite{wang2024videotree,fan2025videoagent}.

To better study how an VLM agent can mimic human mind palace technique, we introduced a novel Video MindPalace Benchmark (VMB) using egocentric video. This is because the moving egocentric view point capture how human explores and exploits in the scene in an untrimmed setting, and therefore closely captured human daily activity is grounded in the 3D world. Our novel benchmark presents a novel set of questions to assess a model's ability to reason about spatial, temporal, and layout-based relationships in real-world environments. VMB requires contextually grounded responses that resemble human understanding. VMB's questions fall into three categories: (1) Enhanced Spatial Localization, requiring precise spatial relations between objects; (2) Contextual Temporal Reasoning, demanding event-based answers reflective of human memory; and (3) Layout-Aware Reasoning, which evaluates a model's grasp of navigation within an environment. These elements together provide a more nuanced and human-aligned assessment of spatio-temporal reasoning.  

\noindent We summarize our contributions as following:
\begin{itemize}
    \item We propose a novel VLM system, VideoMindPalace, which represents long-form video content as a hierarchically-organized graph. VideoMindPalace captures both spatial and temporal aspects, enabling the efficient encoding of key moments dispersed across extensive temporal spans but concentrated in spatial zones. It structures the environment through nodes (spatial concepts like objects, rooms, and activity zones) and edges (spatio-temporal relationships), enhancing LLM comprehension of video content.
    \item We propose the Human-Aligned Spatial and Temporal Reasoning (VMB) benchmark, designed to assess a model's ability to reason about spatial, temporal, and layout-based relationships in real-world environments. VMB challenges models with contextually grounded questions that require human-like reasoning. 
    \item We demonstrate effectiveness across various LVQA benchmarks, including EgoSchema~\cite{mangalam2023egoschema}, NExT-QA~\cite{xiao2021next}, IntentQA~\cite{li2023intentqa}, and Active Memories Benchmark (AMB)~\cite{goletto2025amego} and our newly proposed VMB, addressing complex spatio-temporal reasoning tasks and surpassing the existing methods in these benchmarks.  
\end{itemize}

\section{Related Work}
\label{sec:related}

\noindent\textbf{Structured Video Representation.}
Research on video representations has increasingly emphasized structured approaches, especially by incorporating contextual relationships through graph-based models~\cite{arnab2021unified, baradel2018object, cong2021spatial, brendel2011learning, tsai2019video}. In robotics, scene graphs have shown particular promise for high-level reasoning and planning tasks. Several studies~\cite{gu2024conceptgraphs, wu2024voronav, rajvanshi2024saynav} explore their use in mobile robotics, typically constructing hierarchical scene graphs with layers such as buildings, floors, rooms, and objects. While some works~\cite{rajvanshi2024saynav, agia2022taskography} employ ground truth scene graphs to advance reasoning tasks, others~\cite{liu2023reflect} generate post-execution hierarchical summaries to provide a structured overview of robotic observations. However, these works do not consider how humans interact with objects in video sequences. Another relevant line of work focuses on capturing interactions between objects and actors~\cite{lee-ijcv2015,baradel2018object,arnab2021unified,cong2021spatial} or relationships among actions~\cite{brendel2011learning} to enhance video understanding.
Recent works in egocentric vision have also explored structured video representations. Approaches include grouping video clips by activity threads~\cite{price2022unweavenet} and creating egocentric scene graphs that capture interactions involving the camera wearer~\cite{rodin2024action}. Other methods~\cite{nagarajan2020ego} focus on constructing human-centric representations by mapping spatial relationships in interaction contexts.
While scene graphs in robotics often capture spatial layouts for high-level reasoning and planning, they typically lack human-object interaction details. Additionally, existing works in video representation and egocentric video graphs focus on structured approaches to capture object and actor relationships. However, they lack a design optimized for LLMs. Our method addresses this gap by structuring the graph to be understandable by LLMs, facilitating natural language-based reasoning over video sequences. Another work which is related is AMEGO~\cite{goletto2025amego}, which captures key locations and object interactions in a structured representation but does not incorporate 3D information as we do. Additionally, their representation is purely visual, whereas our structured representation integrates semantic text descriptions, enhancing interpretability and contextual understanding.
\\

\noindent\textbf{Long-Video Understanding with LLMs.}
Long videos have garnered significant interest, with large-scale datasets driving the development of models capable of handling videos lasting from minutes to hours. Recent long video understanding methods~\cite{ren2024timechat, song2024moviechat, maaz2023video, lin2023video, li2024llava, cheng2024videollama, xu2024slowfast} integrate LLMs with video encoders, leveraging their comprehension and generation abilities, often fine-tuned on large-scale labeled video datasets. Others~\cite{zhang2023simple, wang2024videoagent, wang2024videotree} treat long-form video understanding as a natural language question-answering task, first captioning the video and then using LLMs to answer queries. The most relevant works are those leveraging external tools—either LLM agents~\cite{wang2024videoagent, min2024morevqa} for step-by-step executable programs or specialized perception models for each sub-task—to address multimodal tasks without extensive retraining~\cite{wang2024videoagent, wang2024videotree}. VideoTree~\cite{wang2024videotree} addresses this by dynamically extracting keyframes relevant to the query in a coarse-to-fine manner, organizing them in a tree structure where child nodes provide finer details. VideoAgent~\cite{fan2025videoagent} employs a unified memory structure to enhance spatio-temporal reasoning, with two components: (1) temporal memory for text descriptions of short segments and (2) object memory for tracking occurrences of objects and people, enabling LLM performance that rivals or surpasses end-to-end models. Our work follows the similar path of utilizing perception models as tools, but we specifically focus on encoding their outputs in a concise, topologically organized structure that LLMs can readily interpret. Notably, our method generates a hierarchical representation of the entire video, thereby avoiding exhaustive search for different visual queries as in~\cite{wang2024videoagent, wang2024videotree}. 
\\

\begin{figure*}[t]
\centering
\includegraphics[width=0.95\textwidth]{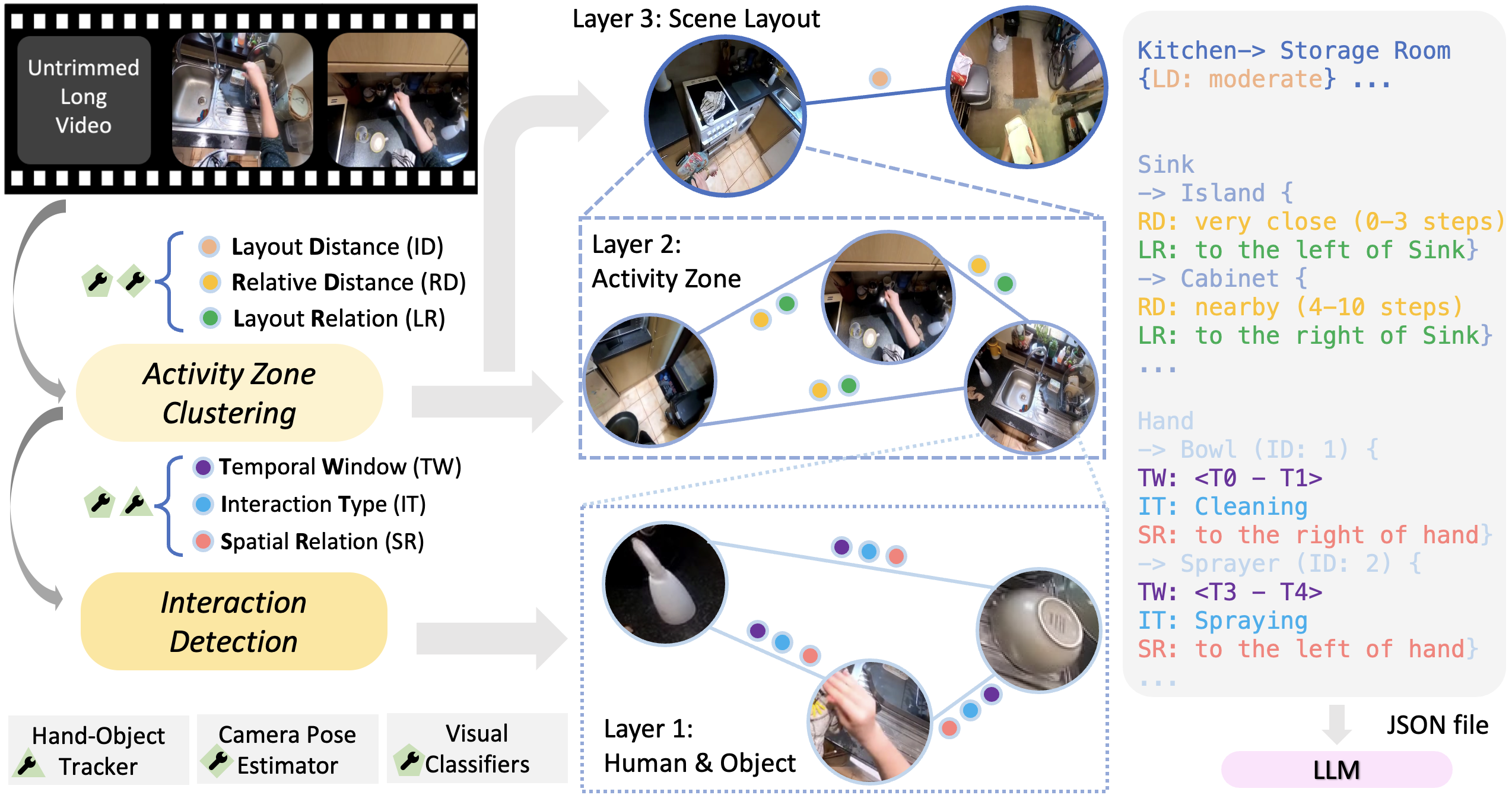}
\vspace{-0.5em}
\caption{Overview of our VideoMindPalace framework. 1) VideoMindPalace is a three-layered graph with nodes representing spatial concepts (e.g., objects, zones, rooms) and edges capturing spatiotemporal relationships. Layer 1 - Human and Object: Nodes represent the human, and detected objects, with edges denoting spatiotemporal connections and interactions. Layer 2 - Activity Zones: Nodes represent specific activity zones with edges showing 3D spatial relationships. Layer 3 - Scene Layout: Nodes represent rooms with edges for relative distances. 2) This graph can be represented in the JSON format used as input to LLMs. The model’s responses are grounded in the physical scene, enabling it to identify locations, locate items of interest, and understand the topological structure of the space.}  
\vspace{-1.5em}
\label{fig:main}
\end{figure*}

\noindent\textbf{Video Understanding Benchmarks.}
Numerous benchmarks assess general video understanding. For instance, NExT-QA~\cite{xiao2021next} includes diverse question types (Temporal, Causal, Descriptive) for short clips, while IntentQA~\cite{li2023intentqa} targets reasoning about human intent in extended videos. Video-language benchmarks like QaEgo4D~\cite{barmann2022did} and EgoSchema~\cite{mangalam2023egoschema} emphasize question-answering tasks in egocentric settings, with models such as GroundVQA~\cite{di2024grounded} developed for this purpose.
A recent benchmark, AMB~\cite{goletto2025amego}, evaluates models on Sequencing, Concurrency, and Temporal Grounding for long egocentric video understanding. In contrast, our VMB benchmark offers both multiple-choice QA and open-ended QA formats, challenging models with advanced reasoning tasks: (1) Enhanced Spatial Localization, requiring precise spatial relationships between objects; (2) Contextual Temporal Reasoning, demanding event-based answers that mimics human memorization; and (3) Layout-Aware Sequential Reasoning, evaluating a model's holistic understanding of spatial and structural information within the scene.

\section{Framework}

Inspired by the Mind Palace technique~\cite{yates2013art}, which involves visualizing physical spaces to mentally store and retrieve key information, we propose a novel representation for long-form video understanding with LLMs - VideoMindPalace, a graph-based structure designed to represent critical moments in long videos by associating them with distinct activity zones. In VideoMindPalace, the video is grounded as a sequence of visits to these zones, with each zone encapsulating a cohesive set of interactions. Benefit from the topological structure, VideoMindPalace can unleash language models' reasoning capabilities to answer questions like: which areas of the scene are most relevant to human actions, and what interactions each zone facilitates.

Our VideoMindPalace is structured as a three-layered graph, with nodes representing spatial concepts (e.g., objects, activity zones, rooms) and edges capturing spatiotemporal relationships, see Fig~\ref{fig:main}.
Formally, we define this hierarchical graph structure as follows:
We construct a three-layer hierarchical graph to represent spatial and interaction relationships across rooms in the scene, activity zones, and entities (humans and objects).

\noindent\textbf{Layer 1 - Human \& Object:} \ In this layer, \( N_1^{(i, j)} = \{ n_1^{(i, j, k)} \mid k \in \{1, \dots, |N_1^{(i, j)}|\} \} \) represents the set of humans and objects in the activity zone, with \( i \) indexing individual room in the scene, \( j \) indexing individual activity zone in each room, \( k \) indexing individual humans or objects, and \( E_1^{(i, j)} \subseteq N_1^{(i, j)} \times N_1^{(i, j)} \) includes edges that encode spatial relationships, interactions, and temporal windows between entities. Hence, the Human \& Object Graph can be defined as \( G_1^{(i, j)} = (N_1^{(i, j)}, E_1^{(i, j)}) \).


\noindent\textbf{Layer 2 - Activity Zones:} \ Here, \( N_2^{(i)} = \{ n_2^{(i, j)} \mid j \in \{1, \dots, |N_2^{(i)}|\} \} \) denotes the set of activity zones within room \( n_1^i \) that the actor explores, such as kitchen sinks, sofas, tables, or cooking counters, and \( E_2^{(i)} \subseteq N_2^{(i)} \times N_2^{(i)} \) captures relative distance and layout relations between activity zones. Each activity zone node \( n_2^{(i, j)} \in N_2^{(i)} \) in the Layer 2 is associated with a Human \& Object subgraph.

\noindent\textbf{Layer 3 - Scene Layout} is represented by a graph \( G_3 = (N_3, E_3) \), where \( N_3 = \{ n_3^i \mid i \in \{3, \dots, |N_3|\} \} \) denotes the set of rooms, and \( E_3 \subseteq N_3 \times N_3 \) represents edges encoding layout distance between rooms. Each room node \( n_3^i \in N_3 \) in the Scene Layout Layer has an associated Actvity Zone subgraph \( G_2^{(i)} = (N_2^{(i)}, E_2^{(i)}) \). 

\noindent The above hierarchical structure enables us to represent and analyze complex spatial and temporal interactions at different levels of granularity.


In the following sections, we first introduce how we leverage multi-object tracking models to predict categories, tracking IDs, and bounding boxes for objects tracked across video frames to construct the Human \& Object Graph in Section~\ref{sec:HOT}. We then present how to discover activity zones for actions from long form videos, using each RGB frame along with its camera pose in Section~\ref{sec:DAZ}. In Section~\ref{sec:GC}, we detail how we use a captioning model to predict semantic labels for activity zones, rooms, and interactions, and then further construct the holistic Scene Layout Graph.

\subsection{Encoding Human-Object Interactions}
\label{sec:HOT}
A key element of our video graph construction pipeline is tracking objects throughout the video and re-identifying previously encountered objects to prevent duplication. Following VideoAgent~\cite{wang2024videoagent}, we integrate the RT-DETR~\cite{zhao2024detrs} object detection model with the ByteTrack~\cite{zhang2022bytetrack} multi-object tracker to generate tracking IDs, categories, and bounding boxes for each detected object across frames. For egocentric videos, which require precise tracking of right and left hands, we use AMEGO~\cite{goletto2025amego} with a human-object interaction tracker to capture interaction tracklets, including the interaction’s start and end frames, bounding box sequences, object ID, and hand side. 

With human-object interaction tracking, we can construct the layer 1 graph \(G_{1} = (N_{1}, E_{1})\) of an activity zone, where each node \(N_{1}\) represents the human (or hands) and any detected objects being manipulated, and edges \(E_{1}\) between these nodes capture interaction types, temporal window, and relative layout. The value of each node is defined as follows: 

\noindent\textbf{Interaction Type:} Each edge is labeled with an interaction type based on the actions inferred by the captioning model (e.g., ``holding," ``cutting," ``placing"). For example, if the human node is associated with a ``sponge," the edge between them might be labeled ``cleaning." 

\noindent\textbf{Spatial Relation:} For human and objects that appear close together in the frame, we define edges to indicate their relative spatial positions (e.g., ``right,'' ``left,''). These relations are determined by analyzing the bounding boxes and spatial arrangement of human and objects in each frame. This provides context about how human and objects are positioned during interactions.

\noindent\textbf{Temporal Window:} Each edge includes a temporal window attribute to capture interaction duration and continuity, defining the time span over which each interaction occurs. These temporal edges enable us to track the evolution of interactions and spatial configurations over time, enhancing the graph’s temporal coherence. 

\subsection{Discovering Activity Zones}
\label{sec:DAZ}
In video analysis, spatial zone discovery is essential for organizing interactions within the scene. Traditional methods, such as visual clustering~\cite{wang2024videotree} and geometric partitioning~\cite{nagarajan2020ego}, often fall short. Visual clustering, for instance, is overly sensitive to specific objects in scenes, so a sink with vegetables versus a sink with bowls may cluster as separate zones. SLAM-based methods can be unreliable due to the rapid camera motions typical of egocentric videos, and is infeasible for exocentric data captured with a static camera.

To address these issues, we implement a zone discovery approach that links views based on both visual content and repeated visits in the long-form video. Rather than relying on a localization network as in Ego-TOPO~\cite{nagarajan2020ego}, we utilize the CLIP model (together with camera pose estimates when available) to identify commonly visited spaces. Frames are considered similar if they meet two criteria: (1) visual similarity and (2) spatial proximity in 3D space.
Using these criteria, we process frames from untrimmed, unlabeled videos to build a topological map. For a video \(\mathcal{V}\) with \(T\) frames \((F_1, \dots, F_T)\) and estimated camera pose position \((x_t, y_t, z_t)\), we construct a graph \(G_{2} = (N_{2}, E_{2})\). Each node \(N_{2}\) represents a spatial zone and contains visits from the video at that location.
We extract visual features for each frame using the CLIP encoder~\cite{radford2021learning}, where \(f_i = CLIP(F_i)\) represents each frame’s visual content. These features provide a compact representation that captures diverse semantics such as scenes and objects. 

The graph initialization begins with a single node \(n_1\) containing the first frame. Each subsequent frame feature \(f_t\) is evaluated for assignment to an existing node based on:
1) Feature-Level Similarity: We calculate the average similarity score \(s_f\) between \(f_t\) and each node \(n \in N\):
   \[
   s_f(f_t, n) = \frac{1}{|n|} \sum_{v \in n} \text{CosineSimilarity}(f_t, f_v) \tag{1}
   \]
   where \(f_v\) is the central frame from each visit in node \(n\). The maximum similarity score is \(s^* = \max_{n \in N} s_f(f_t, n)\).
2) Spatial Proximity: We compute the Euclidean distance \(d\) between the frame’s camera position \((x_t, y_t, z_t)\) and the average position \((x_n, y_n, z_n)\) of each node:
   \[
   d(F_t, n) = \sqrt{(x_t - x_n)^2 + (y_t - y_n)^2 + (z_t - z_n)^2} \tag{2}
   \]
   where \((x_n, y_n, z_n)\) is the mean position of all frames currently assigned to node $n$. The minimum distance across nodes is \(d^* = \min_{n \in N} d(F_t, n)\).
A frame \(F_t\) is assigned to a node \(n^*\) if:
i) feature similarity \(s^*\) exceeds a visual threshold, and
ii) spatial proximity \(d^*\) is below a distance threshold.
If both conditions are met, \(F_t\) is assigned to the most similar and spatially proximate node \(n^*\). If neither condition is met, a new node is created, and an edge connects it to the previous node. To identify the semantic labels of each discovered activity zone and its corresponding room, we feed the center frame of each node into a captioner~\cite{li2023blip}, which predicts both pieces of information. We then group activity zones that belong to the same room accordingly. Additional details can be found in the Supp.

After robustly clustering frames into spatial zones \(N_{2}\), we define edges \(E_{2}\) between these nodes to capture relative distance and spatial layout information. Specifically:

\noindent\textbf{Layout Relation:} Using the camera’s pose data, we define the relative layout of adjacent zones, such as ``left of," ``right of," ``in front of," or ``behind." These descriptors are attached as attributes to the edges, enabling spatial reasoning about how zones are laid out in relation to one another. 

\noindent\textbf{Relative Distance:} To further capture layout, we assign each edge a weight based on the Euclidean distance between nodes in 3D space. This weight reflects the physical proximity of zones, making it possible to infer traversal paths or zones that are often accessed together. 

\noindent The resulting graph represents spatial zones of action (nodes) and 3D layout (edges), allowing the model to focus on key interactions within each zone while disregarding repetitive or irrelevant content.

\subsection{Full Video Graph Construction}

\label{sec:GC}

So far, we discussed how nodes for activity zones and human-object interactions are identified across Layer 1 and Layer 2 and how edges are defined to capture spatial layout, interaction dynamics, and temporal relationships. Each edge type captures a different aspect of the relationships between nodes, enhancing the graph’s expressiveness. For graph \(G_{3} = (N_{3}, E_{3})\) at \textit{Layer 3 - Scene Layout:} nodes \(N_{3}\) represent larger spatial areas, such as ``kitchen" or ``living room," that group multiple activity zones from Layer 2. Edges \(E_{3}\) between room nodes capture relative spatial layout. Nodes for adjacent rooms are connected by edges if they are physically connected between spaces. 

\noindent\textbf{Layout Distance:} This relative distance is determined by analyzing the physical layout and positioning of rooms within the environment, helping to contextualize the spatial flow between areas.

Through this construction, the final graph captures both spatial and temporal relationships across various granularities, from detailed object interactions in Layer 1 to room-level connectivity in Layer 3. This structure allows the graph to represent the environment and actions in a way that supports spatial reasoning, enabling insights into how different zones are organized and used in the environment. The exact implementation details can be found in the Supp.

\section{Video MindPalace Benchmark}

The \textbf{Video MindPalace Benchmark (VMB)} introduces a suite of questions designed to evaluate a model's understanding of spatial, temporal, and layout-based reasoning in real-world environments. VMB requires a model to interpret contextually grounded answers that align closely with human responses in natural settings. VMB's multiple-choice QA set focuses on three primary question types: 

\noindent\textbf{Enhanced Spatial Localization:} Traditional benchmarks might respond to ``Where is my key?'' with a simple location (``on the table''). VMB pushes further by requiring a model to specify spatial relations with nearby objects (``on the table, to the right of a book'') adding a nuanced understanding of object context and layout. 

\noindent\textbf{Contextual Temporal Reasoning:} While most benchmarks provide time-based answers for temporal queries (e.g.,``from 10s to 50s in the video''), VMB emphasizes event-based responses that mirror human-like recall. For example, to answer ``When did I use my pencil?'' the model should respond ``right after you opened your laptop'', reflecting real-life associative memory. 

\noindent\textbf{Layout-Aware Reasoning:} VMB introduces layout reasoning questions that assess a model’s understanding of navigation within an environment. For instance, when asked, ``Is there any location situated between the table and the refrigerator?'' the model must infer spatial relationships, identifying objects or areas based on their arrangement and proximity within the setting, to answer accurately (e.g., ``a dining chair, slightly to the left of the table''). 

For the open-ended QA set, we test the model's ability to describe fine-grained actions and transitions within the scene, including questions that assess spatial sequencing and navigation understanding. For example, when asked, ``What did the human do after washing the potato?'' the model must articulate actions and transitions (e.g., ``\textit{turned right} to the cooking bench to heat the frying pan, then \textit{took a few steps left} to the shelf to prepare a clean bowl''). This level of response requires a coherent mental map of the environment and an understanding of activity flow.

The benchmark construction process involves generating and validating both multiple-choice and open-ended questions that require grounded understanding of the environment captured in the video. We construct our benchmark using videos from the EPIC-KITCHENS and Ego-4D datasets, which consist of lengthy, egocentric recordings of participants performing daily activities in a kitchen environment. For each multiple-choice set, covering question types such as Spatial, Temporal, and Layout, we categorize video lengths into Short (less than 3 minutes), Medium (3 to 10 minutes), and Long (over 10 minutes), with approximately 600 questions for each length category. Additionally, the open-ended question set includes 30 questions prompting the model to describe the actions and transitions of the camera wearer within the explored environment, requiring fine-grained answers. More details can be found in the Supp.

\begin{table}[t]
\centering
\small
\begin{tabular}{lccc}
\toprule
\textbf{Method} & \textbf{NExTQA} & \textbf{EgoSchema} & \textbf{IntentQA} \\
\midrule
Video-LLaMA2~\cite{cheng2024videollama}   & - & 53.3 & - \\
SF-LLAVA-34B~\cite{xu2024slowfast}  & 72.0 & 55.8 & 66.5 \\
LLoVi~\cite{zhang2023simple}   & 67.7 & 61.2 & 64.0 \\
VideoAgent~\cite{wang2024videoagent}  & 71.3 & 60.2 & - \\
VideoTree~\cite{wang2024videotree}   & 73.5 & 66.2 & 66.9 \\
IG-VLM~\cite{kim2024image}    & 68.6 & 59.8 & 64.2 \\
\rowcolor[HTML]{cef8d1} Ours   & \textbf{75.8} & \textbf{68.6} & \textbf{70.1} \\
\bottomrule
\end{tabular}
\vspace{-0.5em}
\caption{We compare VideoMindPalace with other supervised fine-tuning and training-free methods on EgoSchema~\cite{mangalam2023egoschema}, NExT-QA~\cite{xiao2021next} and IntentQA~\cite{li2023intentqa}, and demonstrate the effectiveness of VideoMindPalace against both types.}
\vspace{-1.5em}
\label{tab:MC_benchmark}
\end{table}

\begin{table*}[t]
\centering
\small
\begin{tabular}{lccccccccccccc}
\toprule
\textbf{Method} & \multicolumn{4}{c}{\textbf{Spatial}} & \multicolumn{4}{c}{\textbf{Temporal}} & \multicolumn{4}{c}{\textbf{Layout}} & \multicolumn{1}{c}{\textbf{OE}}\\
\cmidrule(lr){2-5} \cmidrule(lr){6-9} \cmidrule(lr){10-13}
& \textbf{S} & \textbf{M} & \textbf{L} & \textbf{Avg} & \textbf{S} & \textbf{M} & \textbf{L} & \textbf{Avg} & \textbf{S} & \textbf{M} & \textbf{L} & \textbf{Avg} & \multicolumn{1}{c}{(0-5)} \\
\midrule
Random & 20.0 & 20.0 & 20.0 & 20.0 & 20.0 & 20.0 & 20.0 & 20.0 & 20.0 & 20.0 & 20.0 & 20.0&  - \\
IG\_VLM   & 32.8 & 31.2 & 30.5 & 31.5 & 36.2 & 33.8 & 30.5 & 33.5 & 22.4 & 21.9 & 21.1 & 21.8 & 1.1   \\
LLoVi    & 44.5 & 41.8 & 40.6 & 42.3 & 55.2 & 53.1 & 50.1 & 52.8 & 23.1 & 22.3 & 21.8 & 22.4 & 3.3  \\
VideoTree & 43.2 & 40.8 & 37.5 & 40.5 & 47.0 & 45.5 & 43.4 & 45.3& 23.9 & 22.6 & 22.2 & 22.9 & 2.5   \\
VideoAgent & 50.5 & 45.9 & 44.7 & 47.0 & 58.2 & 55.3 & 51.2 & 54.9 & 25.0 & 24.7 & 23.8 & 24.5 & 2.7   \\
\rowcolor[HTML]{cef8d1} Ours & 53.0 & 51.5 & 49.1 & \textbf{51.2} & 57.2 & 56.6 & 55.1 & \textbf{56.3} & 37.0 & 33.7 & 32.8 & \textbf{34.5} & \textbf{3.8}  \\
\bottomrule
\end{tabular}
\vspace{-0.5em}
\caption{We compare VideoMindPalace with IG\_VLM~\cite{kim2024image}, LLoVi~\cite{zhang2023simple}, VideoTree~\cite{wang2024videotree} and VideoAgent~\cite{fan2025videoagent} on VMB benchmark evaluated across different video lengths and question types. Video lengths are categorized as Short (less than 3 minutes), Medium (3 to 10 minutes), and Long (over 10 minutes). Our VideoMindPalace model achieves the best performance on averaged performance across all video length categories. OE stands for Open-Ended.}
\vspace{-1.5em}
\label{tab:vmb}
\end{table*}

\section{Experiments}


\noindent\textbf{Benchmarks and Metrics}.\ Multiple Choice VideoQA presents a set of multiple choice options to Video LLMs and evaluates their capability of picking the correct choice. Specifically, we evaluate our model on NExTQA~\cite{xiao2021next}, EgoSchema~\cite{mangalam2023egoschema} validation set, IntentQA~\cite{li2023intentqa}, AMB~\cite{goletto2025amego} and Multiple choice QA set of our VMB. The accuracy of selecting the correct option is used as the evaluation metric. Open-Ended VideoQA expects the model to generate answers in freeform in response to a question for a video. We evaluate this task using the open-ended QA set of our VMB. We use the LLM-assisted evaluation to assess the quality (score ranging from 0 to 5) of the models.

\noindent\textbf{Implementation Details}.\ We adopt GPT-4~\cite{achiam2023gpt} as our language model (LLM) for generating all primary results, including ours and other methods. Following the approaches in ~\cite{fan2025videoagent, goletto2025amego}, we use ByteTracker~\cite{zhang2022bytetrack} and EgoSTARK~\cite{tang2024egotracks} as our primary tracking algorithms. To predict semantic labels for both locations and interactions, we follow~\cite{goletto2025amego} to utilize captioner BLIP-2~\cite{li2023blip}. Additionally, we use CLIP ViT-B to encode the visual features of key frames and retrieve camera poses  using ~\cite{tschernezki2024epic}.

\subsection{Main Results}


\noindent\textbf{Standard VQA.} Similar to previous works~\cite{wang2024videotree, zhang2023simple}, we evaluate our approach on three benchmark datasets for standard video question answering: EgoSchema, IntentQA, and NExT-QA. These benchmarks are designed to assess causal and temporal reasoning, general scene understanding, and question types such as ``Why?" and ``How?".
Table~\ref{tab:MC_benchmark} shows that our method outperforms other models across multiple-choice VideoQA benchmarks, achieving the highest accuracy on NExT-QA (75.8\%), EgoSchema (68.6\%), and IntentQA (70.1\%). Notably, it surpasses the next-best model, VideoTree, by 2.4\% on EgoSchema and 3.2\% on IntentQA, demonstrating strong performance in general reasoning, temporal understanding, and answering ``Why?" and ``How?" questions in video.

\noindent\textbf{VMB Benchmark.} The results in Table~\ref{tab:vmb} demonstrate that our VideoMindPalace model outperforms other methods across various video lengths (Short, Medium, and Long) and question types (Spatial, Temporal, Layout, and Open-Ended). Specifically, VideoMindPalace achieves the highest average scores in the Spatial (51.2) and Temporal (56.3) categories, indicating strong performance in capturing both spatial relationships and temporal coherence within videos. It also leads in the Layout category with an average score of 34.5, showing its capability in understanding spatial arrangements. Additionally, in the Open-Ended category, VideoMindPalace attains the highest score of 3.8, suggesting better overall comprehension in answering complex questions. Notably, VideoMindPalace performs especially well on Long videos, highlighting its effectiveness in handling extended temporal contexts, and maintains consistent top performance across all video lengths.

\begin{table*}[t]
\small
\centering
\begin{tabular}{l p{1.0cm} p{1.0cm} p{1.0cm} p{1.0cm} p{1.0cm} p{1.0cm} p{1.0cm} p{1.0cm} p{1.0cm} p{1.0cm} p{1.0cm} p{1.0cm} c}
\toprule
\textbf{Method} & \multicolumn{4}{c}{\textcolor{blue}{SQ}} & \multicolumn{2}{c}{\textcolor{red}{CO}} & \multicolumn{2}{c}{\textcolor{green}{TG}} & \textbf{Total} \\
\cmidrule(lr){2-5} \cmidrule(lr){6-7} \cmidrule(lr){8-9}
& \textbf{Q1} & \textbf{Q2} & \textbf{Q3} & \textbf{Q4} & \textbf{Q5} & \textbf{Q6} & \textbf{Q7} & \textbf{Q8} & \textbf{} \\
\midrule
Random & 20.0 & 20.0 & 20.0 & 20.0 & 20.0 & 20.0 & 20.0 & 20.0 & 20.0 \\
\midrule
S-QA  & 23.9 & 22.0 & 22.5 & 23.3 & 27.5 & 27.0 & 20.2 & 24.1 & 23.6 \\
LLoVi  & 22.3 & 21.4 & 21.8 & 22.2 & 25.6 & 26.7 & 18.1 & 22.2 & 22.4 \\
AMEGO  & 33.7 & 36.3 & 37.2 & 38.3 & 27.6 & 44.3 & 34.7 & 48.9 & 36.3 \\
IG-VLM  & 17.3 & 23.0 & 21.4 & 31.6 & 33.1 & 38.4 & 18.3 & 19.6 &  25.3\\
VideoTree  & 17.5 & 25.3 & 25.0 & 32.7 & 34.3 & 42.5 &15.4  & 22.9 &  27.0\\
VideoAgent  & 35.6 & 26.1 & 23.3 & 34.2 & 38.2 & 39.2 & 38.2 & 43.5 & 34.6 \\
\rowcolor[HTML]{cef8d1} Ours & 44.7 & 27.9 & 24.2 & 34.5 & 42.7 & 46.4 & 41.0 & 57.4 & \textbf{37.8} \\
\bottomrule
\end{tabular}
\vspace{-0.5em}
\caption{We compare VideoMindPalace with S-QA~\cite{radford2021learning}, LLoVi~\cite{zhang2023simple}, AMEGO~\cite{goletto2025amego}, IG-VLM~\cite{kim2024image}  on Active Memory Benchmark~\cite{goletto2025amego}, VideoTree~\cite{wang2024videotree} and VideoAgent~\cite{fan2025videoagent} on Active Memory Benchmark~\cite{goletto2025amego}, including three categories across questions Q1 to Q8, with overall Total accuracy. Sequencing (SQ) questions assess the ability to identify the order of events. Concurrency (CO) questions evaluate recognition of simultaneous interactions. Temporal grounding (TG) questions test the model’s ability to retrieve all intervals of interactions with an active object or location within a long video.}
\label{tab:amego}
\end{table*}

\noindent\textbf{Active Memories Benchmark.}  AMB~\cite{goletto2025amego} studies interactions between active objects, locations, and their interplay in long egocentric videos, essential for understanding daily human activity. The benchmark includes 20.5k multiple-choice queries across various reasoning levels, from simple object-use questions (e.g., “What did I use with [VQ]?” where [VQ] is a visual crop of an object) to complex object-location interactions (e.g., “Where did I use [VQ]?”). Given a set of visual answers [VA], the task involves selecting the correct object or location representation based on the interaction. Each query and answer (visual query [VQ], visual object answer [VA], location query [LQ], and location answer [LA]) is represented as a visual crop, allowing models to rely purely on visual data for responses. Since many training-free approaches rely on text-based video representations, we first use an off-the-shelf captioner to convert visual queries into semantic descriptions, enabling us to evaluate these approaches on the benchmark effectively.
Table~\ref{tab:amego} summarizes the performance of various methods on three VideoQA reasoning tasks: Sequencing (SQ), Concurrency (CO), and Temporal Grounding (TG). Our framework achieves the highest total accuracy (37.8\%), outperforming all other models. Specifically, our model excels in Temporal Grounding with an accuracy of 41.0\% and 57.4\%, indicating strong capability in associating specific temporal moments with activity zones. However, our method lags behind AMEGO on certain spatial grounding tasks, due to the error (e.g. hand side prediction) when converting visual query to semantic description. This is a pitfall shared by all methods that using text-only LLM for visual query retrieval.

\begin{figure*}[t]
    \centering
    \includegraphics[width=1.0\textwidth]{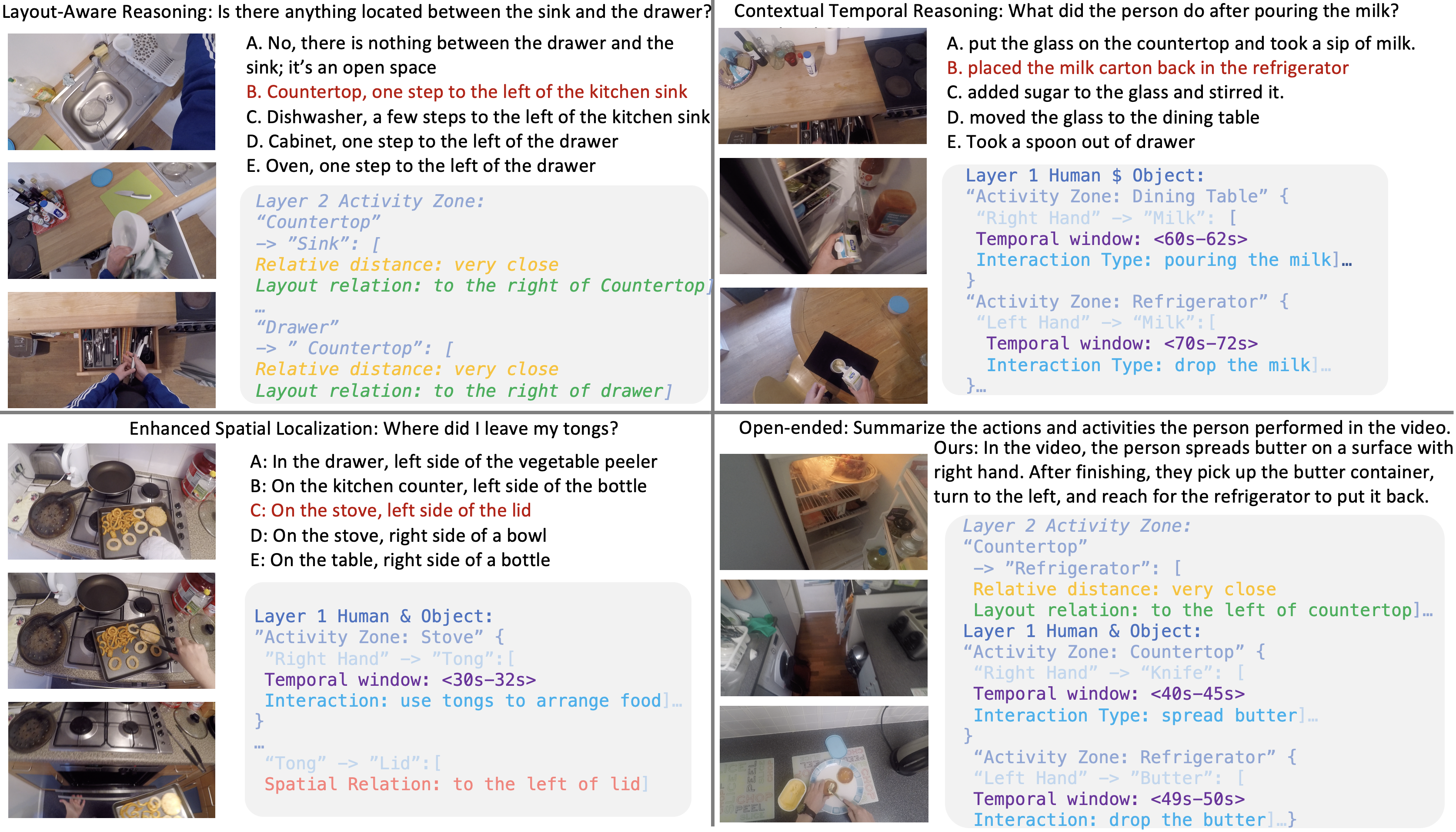}
    \vspace{-1.5em}
    \caption{Qualitative results of VideoMindPalace on the VMB benchmark, with an example for each question type. To explore how VideoMindPalace successfully answers these questions, we prompt GPT-4 to identify the specific parts of the graph that provide sufficient information to answer each question accurately.}
    \label{fig:results}
    \vspace{-1.5em}
\end{figure*} 


\subsection{Visualization and Discussion}
Qualitative results in Figure~\ref{fig:results} reveal VideoMindPalace's reasoning capabilities across different question types. For each question type, we present a representative example. To explore how VideoMindPalace successfully answers these questions, we prompt GPT-4 to identify the specific parts of the graph that contain sufficient information for accurate inference. (Refer to Supp. for the exact prompt we used.) 

As shown in Figure~\ref{fig:results}, when answering a question that involves determining whether an object or person is located between two locations, VideoMindPalace leverages nodes and edges within the Activity Zone Layer that capture relative distance and layout relationships. This allows the model to identify intermediary zones or objects accurately. Similarly, for questions regarding temporal sequences (e.g., ``What happened before or after an event?"), VideoMindPalace references edges within the Human and Object Layer to retrieve interaction relationships along with temporal windows. By isolating these key structural components, VideoMindPalace can align the graph's representation with the spatial or temporal reasoning needed to answer each question type correctly. We also provide detailed ablation studis and additional analysis of our Video MindPalace system in Supp.

\section{Limitation and Conclusion}


\noindent\textbf{Limitations}.\ Our VideoMindPalace framework utilizes perception models to construct the graph; however, errors introduced by these models can propagate through to the final graph, potentially impacting the accuracy of the representation. Additionally, the current version of VideoMindPalace does not encode low-level visual details of objects, such as texture or color. Consequently, it may be unable to answer questions that rely on such specifics—such as ``Where did I use the red mug?''—if multiple mugs of different colors are present in the video.


\noindent\textbf{Conclusion}.\
In conclusion, VideoMindPalace introduces a new paradigm for long-form video understanding by adopting a mind-palace-inspired approach that structures complex, dispersed events into a cohesive, topologically organized semantic graph. By grounding temporally distant yet spatially relevant moments within a layered graph representation, VideoMindPalace achieves a nuanced understanding that overcomes traditional challenges of information overload and redundancy inherent in long video analysis. Future work could focus on refining the graph construction process by enhancing the distillation of knowledge from various perceptual sources, ultimately creating a more intuitive and comprehensible final graph. 

\bibliography{ref}
\bibliographystyle{abbrvnat}

\newpage 


\clearpage

In this appendix, we report additional details on VideoMindPalace, the Video MindPalace Benchmark (VMB), additional results and visualizations. In section~\ref{sec:More_vis}, we include more visualizations of both queries and qualitative results across different kinds of questions in VMB. We then give more information on the graph construction for VideoMindPalace and how we prompt VideoMindPalace to obtain the answers for the questions in VMB in section~\ref{sec:More_method}. Next, in section~\ref{sec:More_res}, we present additional ablations for VideoMindPalace. Finally, we further detail VMB in section~\ref{sec:More_bench}.

\section{More qualitative results}
\label{sec:More_vis}

\begin{figure*}[t]
    \centering
    \includegraphics[width=1.0\textwidth]{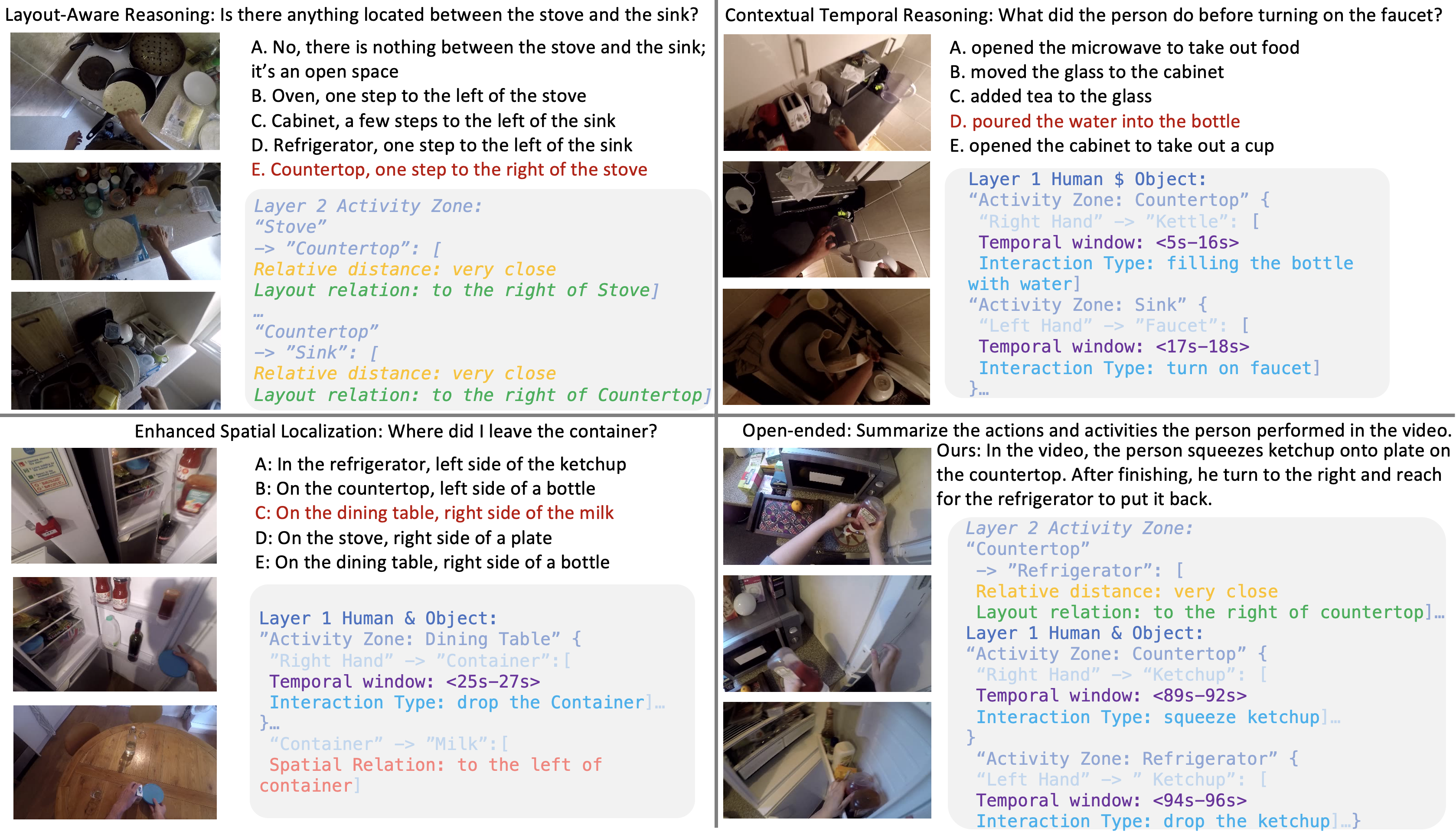}
    \vspace{-1.5em}
    \caption{More qualitative results of VideoMindPalace on the VMB benchmark, showcasing examples for each question type. To demonstrate how VideoMindPalace effectively answers these questions, we leverage GPT-4 to pinpoint specific graph components that provide the necessary information for accurate responses.}
    \label{fig:more_vis_res}
    \vspace{-1.5em}
\end{figure*}

In the figure~\ref{fig:more_vis_res}, we showcase additional examples of VideoMindPalace's reasoning capabilities across various question types. For each question type, we provide a representative example. To illustrate how VideoMindPalace effectively answer these questions, we utilize GPT-4 to identify specific segments of the graph containing the necessary information for accurate inference using the following prompt: ``You are an expert in analyzing semantic graphs generated from video content. In the following graph, nodes capture spatial concepts (e.g., objects, activity zones, rooms), and edges signify spatiotemporal, layout relationships and human-object interaction. Given a query, your task is to identify specific segments of the graph that provide sufficient information to answer the query accurately. Input: 1. Constructed graph: [...], 2. Query and options: [...].''

\section{More graph construction details}
\label{sec:More_method}

\subsection{Prompt design of zero-shot video QA }
We employ the following prompts to extract critical information from the constructed graph for zero-shot video question answering tasks: 
1. For Multiple-Choice Video QA: ``You are an expert at reasoning with semantic graphs to analyze video content. In the following graph, nodes capture spatial concepts (e.g., objects, activity zones, rooms), and edges signify spatiotemporal, layout relationships and human-object interaction. Your task is to answer a query based on the provided graph by selecting the correct answer from given options. Input: 1. Constructed graph: [...], 2. Query and options: [...]. Task: Analyze the graph and determine which option correctly answers the query. Provide only the correct answer (e.g., "A") without additional explanations.''. 2. For open-ended video QA: ``You are an expert in analyzing video content to summarize human actions and activities. In the following graph, nodes capture spatial concepts (e.g., objects, activity zones, rooms), and edges signify spatiotemporal, layout relationships and human-object interaction. Your task is to extract and describe fine-grained actions, transitions, and spatial sequences performed by a person in the video. The summary should provide a coherent and detailed account of the activity flow, highlighting both actions and their relationships to the environment. Input: Constructed graph: [...]. ''.

\subsection{Graph construction heuristic}
Layer 1 - Human and Object: 
we consider \(u\) and \(v\) represent two entities in the scene, such as a human or an object. Each detected object is assigned a unique identifier, denoted as \( \text{ID}_u \), and categorized with a semantic label \( C_u \). The spatial location of an object in frame \( t \) is represented by its bounding box \( B_u(t) = (x_{\text{min}}, y_{\text{min}}, x_{\text{max}}, y_{\text{max}}) \). Between the two nodes \(N_u\) and \(N_v\), three types of edges are defined, each capturing a specific relationship. 
The first type, the Interaction Type Edge (\(I_{uv}\)), encodes the predicted interaction type between \(u\) and \(v\), as inferred by a captioning model BLIP-2~\cite{li2023blip} (e.g., \textit{"cutting," "holding,"} or \textit{"placing"}). 
The second type, the Temporal Window Edge (\(T_{uv}\)), specifies the duration of the interaction as \(T_{uv} = (t_{\text{start}}, t_{\text{end}})\), where \(t_{\text{start}} = \frac{f_{\text{start}}}{\text{FPS}}\) and \(t_{\text{end}} = \frac{f_{\text{end}}}{\text{FPS}}\). Here, \(f_{\text{start}}\) and \(f_{\text{end}}\) represent the start and end frames of the interaction, respectively, as predicted by a human-object interaction tracker used in ~\cite{goletto2025amego}.
The third type, the Spatial Relationship Edge (\(S_{uv}\)), captures the relative position of \(u\) with respect to \(v\). Using the bounding box centers at \(t_{\text{start}}\), the relative horizontal position is determined based on the \(x_{\text{center}}\) coordinates, where \(x_{\text{center}} = \frac{x_{\text{min}} + x_{\text{max}}}{2}\). If \(x_{\text{center}, u} > x_{\text{center}, v}\), \(u\) is labeled as \textit{"right"}; otherwise, \(u\) is labeled as \textit{"left"}. 
Together, these edges encode the semantic, spatial, and temporal relationships between the nodes, forming a comprehensive graph.

Layer 2 - Activity Zones:
To estimate layout relations, we first consider two activity zones, which is clustered using method introduced at~\ref{sec:DAZ}, \(A\) and \(B\), derived from the video data. Using the centroid positions \((x_A, y_A, z_A)\) and \((x_B, y_B, z_B)\) of each zone, we define their relative layout based on their spatial arrangement. For instance, \(B\) is labeled as "left of" \(A\) if \(x_B > x_A\) and the differences along \(y\) and \(z\) axes are within a predefined threshold of 0.5. Similarly, "right of," "in front of," and "behind" relationships are determined by comparing coordinate displacements along \(x\), \(y\), and \(z\) axes. These directional descriptors are attached as attributes to the edge connecting the zones. To capture relative distances, we compute the Euclidean distance between the centroids of \(A\) and \(B\) as \(d_{AB} = \sqrt{(x_B - x_A)^2 + (y_B - y_A)^2 + (z_B - z_A)^2}\). Additionally, we normalize the distance for scale invariance as \(d_{AB}^\text{normalized} = \frac{d_{AB} - d_\text{min}}{d_\text{max} - d_\text{min}}\), where \(d_\text{min}\) and \(d_\text{max}\) are the minimum and maximum distances across all zone pairs. we categorize the distances into five levels based on thresholds: \(0.1, 0.2, 0.4, 0.7\). Distances falling below these thresholds are labeled as \textit{very close} (0--3 steps), \textit{close} (4--6 steps), \textit{moderate} (7--10 steps), \textit{far} (11--15 steps), or \textit{very far} (16+ steps). This categorization allows us to qualitatively assess physical proximity between zones. This combined representation, with directional relationships and normalized distances, enables reasoning about both the spatial layout and physical proximity of the two activity zones.

Layer 3 - Scene Layout:
To construct Layout Distance at Layer 3, we group multiple activity zones from Layer 2 into higher-level spatial groupings based on room-level context using captions generated by the BLIP-2 model. These captions help identify the specific room where each activity zone belongs, such as "kitchen" or "living room." The layout distance between rooms is determined by analyzing their physical layout and positioning within the environment. Consider two rooms, \(R_1\) and \(R_2\). Using the centroid coordinates \((x_r, y_r, z_r)\) of each room—calculated as the average of the centers of the activity zones belonging to that room—we compute the Euclidean distance between them as \(d_{R_1R_2} = \sqrt{(x_{R_2} - x_{R_1})^2 + (y_{R_2} - y_{R_1})^2 + (z_{R_2} - z_{R_1})^2}\). This distance estimation follows the same approach used in Layer 2 for determining layout distances. These room-level distances contextualize the spatial flow between areas, capturing both proximity and potential movement pathways. 

Lastly, as mentioned in Section~\ref{sec:DAZ}, two additional hyperparameters are used: the visual threshold for feature similarity, \(s^* = 0.6\), and the distance threshold for spatial proximity, \(d^* = 0.5\).

\section{Ablation Studies}
\label{sec:More_res}

\paragraph{Cluster by location vs Split by temporal window}
We hypothesize that clustering temporally distant yet spatially relevant frames within a layered graph enables VideoMindPalace to reduce information overload and redundancy in long video analysis. To validate this, we compare our approach with an alternative method that segments the video into shorter chunks and builds separate graphs for each chunk, setting the number of chunks equal to our number of clusters for a direct comparison. Table~\ref{tab:compare_two}
shows that our location-based clustering method outperforms temporal window segmentation across all video lengths, with improvements especially notable in Medium and Long videos (2.7\% and 3.4\% higher, respectively). This indicates that clustering by location effectively reduces redundancy while preserving spatial relevance in longer videos.

\begin{table}[h]
\centering
\small
\begin{tabular}{lccc}
\toprule
Method & Short & Medium & Long \\
\midrule
Split by temporal window  & 48.5 & 44.5 & 42.2 \\
Cluster by location       & 49.1 & 47.2 & 45.6 \\
\bottomrule
\end{tabular}
\vspace{-0.5em}
\caption{Comparison of performance (\%) between temporal window segmentation and location-based clustering across different video lengths (Short, Medium, and Long).}
\vspace{-1.5em}
\label{tab:compare_two}
\end{table}

\paragraph{Impact of different LLMs on reasoning performance} To validate the effectiveness of our method in improving reading performance across various reasoning tasks, we conducted experiments with different sizes of large language models and compared the results with LLoVi. Table~\ref{tab:llm_reasoning_performance} highlights the performance of both methods using GPT-3.5 and GPT-4 on spatial, temporal, and layout reasoning tasks in VMB dataset. These results demonstrate that our method not only leverages the increased capacity of GPT-4 effectively but also exhibits greater robustness to variations in LLM sizes compared to LLoVi. For instance, in the spatial and temporal tasks, the performance gap between GPT-3.5 and GPT-4 for our method is 4.5\% and 5.6\%, respectively, whereas the gap for LLoVi is significantly larger at 6.2\% and 7.8\%. In the layout task, since LLoVi's performance remains close to random guessing for both GPT-3.5 and GPT-4, a direct comparison with our method is less meaningful. We hypothesize that this improvement stems from our graph-based representation, which provides a more structured and intuitive representation that is easier for LLMs to process and comprehend compared to the unstructured representation used in LLoVi.

\begin{table}[h]
\centering
\small
\begin{tabular}{lcccc}
\toprule
Method & LLM & Spatial & Temporal & Layout \\
\midrule
LLoVi      & GPT-3.5 & 36.1 & 45.0 & 21.2 \\
Ours      & GPT-3.5 & 46.7 & 50.6 & 31.0 \\
\hline
LLoVi      & GPT-4 & 42.3 & 52.8 & 22.4 \\
Ours      & GPT-4 & 51.2 & 56.3 & 34.5 \\
\bottomrule
\end{tabular}
\vspace{-0.5em}
\caption{Performance comparison of LLoVi and our method across spatial, temporal, and layout reasoning tasks in VMB using GPT-3.5 and GPT-4.}
\vspace{-1.5em}
\label{tab:llm_reasoning_performance}
\end{table}

\section{Video Mind Palace Benchmark}
\label{sec:More_bench}

\subsection{Query creation}

To construct a robust set of VideoQA queries for the proposed Video MindPalace Benchmark, we employed a systematic pipeline combining LLM-generated questions with human verification to ensure quality and consistency. From each video, we first extract keyframes sampled at 1 frame per second (1 fps) to succinctly represent the content. For each keyframe, we generate a descriptive caption using GPT-4 and provide a detailed textual description of detected objects, including their IDs and bounding box coordinates. These inputs are fed into GPT-4 to generate diverse reasoning questions requiring spatial, temporal, and layout-aware understanding, such as: "Which object is to the left of the dining table?" (spatial), "What event happens immediately after the person enters the kitchen?" (temporal), and "How are the sofa and coffee table arranged in relation to the TV?" (layout-aware). Each question is accompanied by five answer options, including one correct answer and four distractors designed to challenge reasoning abilities. For open-ended queries, we prompt GPT-4 to generate detailed captions for each keyframe and compile these captions to summarize the actions and activities performed by the individual throughout the video. To ensure accuracy, all questions and answers undergo rigorous review by human annotators, who validate correctness, correct errors, and refine phrasing for clarity and consistency. This human-in-the-loop step is critical to maintaining high-quality questions across the benchmark. The finalized questions are then tagged with metadata, such as reasoning type, video length, and associated video segments, and compiled into the Video MindPalace Benchmark. This process ensures a comprehensive and reliable set of VideoQA queries for evaluating video understanding models.

\begin{figure}[h]
    \centering
    \includegraphics[width=0.4\textwidth]{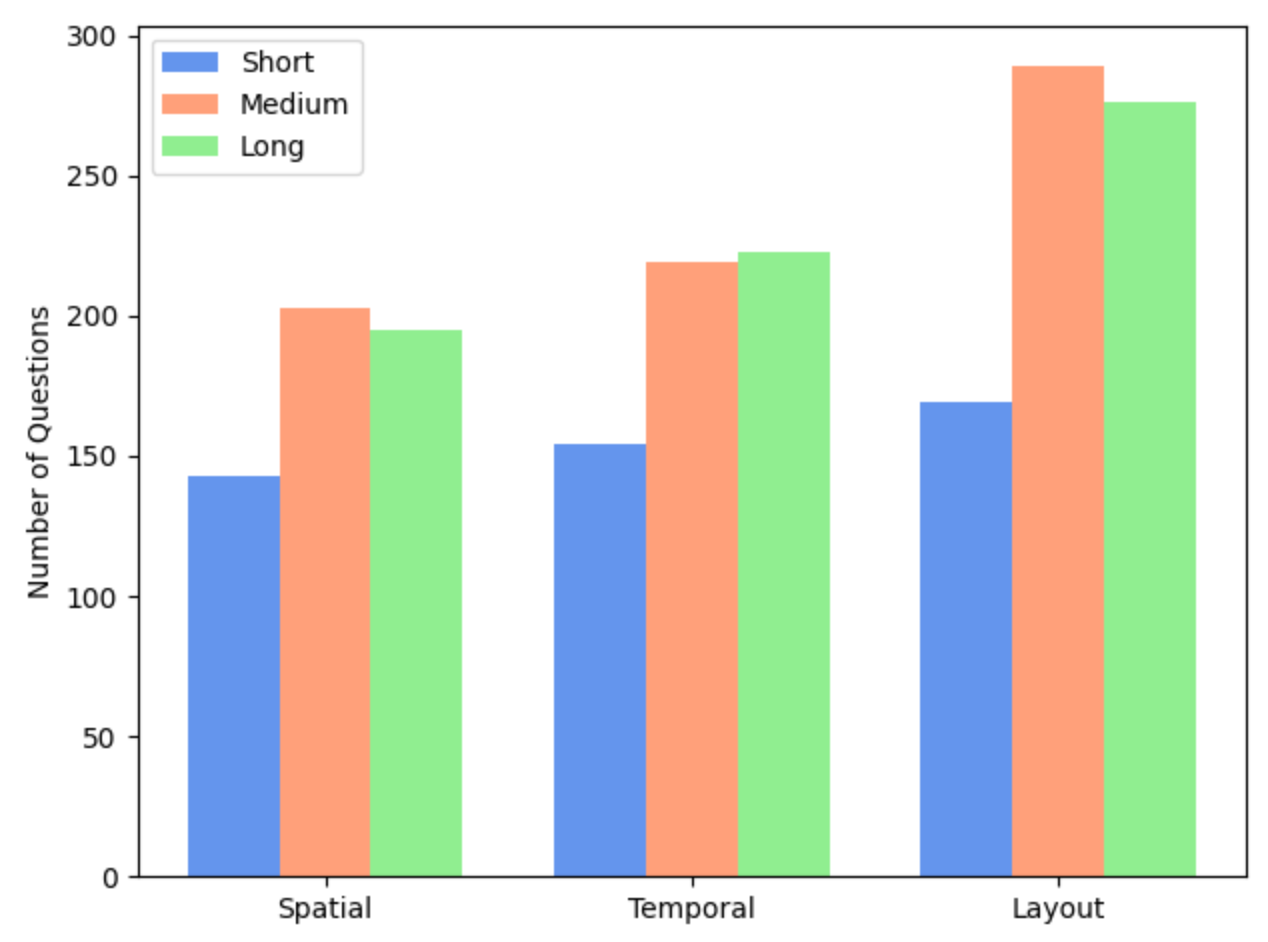}
    \caption{Query Distribution by Video length and Reasoning Categories.}
    \label{fig:dataset_stats}
    \vspace{-1.5em}
\end{figure} 

\subsection{Statistics}
We construct our benchmark using 200 videos sourced from the EPIC-KITCHENS and Ego-4D datasets, both of which consist of long, unscripted egocentric recordings capturing participants performing daily activities in various environments. On average, the selected videos are 11 minutes in length. Specifically, 68 videos are categorized as short (less than 3 minutes), 85 videos as medium-length (3 to 10 minutes), 35 as long (10 to 30 minutes), and 12 as very long (over 30 minutes). Each video length category includes between 100 and 300 questions, spanning all three types of queries, resulting in a total of approximately 1,800 questions in the benchmark. For a detailed distribution of query types by video duration and reasoning category, please refer to Fig~\ref{fig:dataset_stats}.

\end{document}